\documentclass[10pt,twocolumn,letterpaper]{article}

\usepackage{iccv}
\usepackage{times}
\usepackage{epsfig}
\usepackage{graphicx}
\usepackage{amsmath}
\usepackage{amssymb}

\usepackage{pifont}
\usepackage{booktabs}
\usepackage{multirow}
\usepackage{capt-of}
\usepackage{makecell}
\usepackage{colortbl}
\usepackage{dsfont}
\usepackage{comment}
\usepackage[dvipsnames]{xcolor}

\usepackage[accsupp]{axessibility}  %

\usepackage{tocloft}

\makeatletter
\renewcommand{\numberline}[1]{%
  \@cftbsnum #1\@cftasnum\hspace*{1em}\@cftasnumb%
}
\makeatother

\usepackage[pagebackref=true,breaklinks=true,colorlinks,bookmarks=false]{hyperref}

\usepackage[capitalize]{cleveref}
\crefname{section}{Sec.}{Secs.}
\Crefname{section}{Section}{Sections}
\Crefname{table}{Table}{Tables}
\crefname{table}{Tab.}{Tabs.}

\renewcommand{\paragraph}[1]{\vspace{1.25mm}\noindent\textbf{#1}}

\definecolor{baselinecolor}{gray}{.9}
\definecolor{darkgreen}{rgb}{0.13, 0.55, 0.13}

\let\originalleft\left
\let\originalright\right
\renewcommand{\left}{\mathopen{}\mathclose\bgroup\originalleft}
\renewcommand{\right}{\aftergroup\egroup\originalright}

\iccvfinalcopy %

\def\iccvPaperID{5097} %

\begin{document}

\title{Beyond Known Clusters: Probe New Prototypes for Efficient Generalized Class Discovery}

\author{%
  Ye Wang\textsuperscript{$1$} \qquad
  Yaxiong Wang\textsuperscript{$2$} \qquad
  Yujiao Wu\textsuperscript{$3$} \qquad
  Bingchen Zhao\textsuperscript{$4$} \qquad
  Xueming Qian\textsuperscript{$1$} \\
  \textsuperscript{$1$}Xi'an Jiaotong University \quad
  \textsuperscript{$2$}Hefei University of Technology \quad
  \textsuperscript{$3$}CSIRO \\
  \textsuperscript{$4$}University of Edinburgh \\
  \tt\small \{xjtu2wangye@stu,wangyx15@stu,qianxm@mail\}.xjtu.edu.cn \\
  \tt\small yujiaowu111@gmial.com \qquad
  \tt bingchen.zhao@ed.ac.uk \\
}



\maketitle


\begin{abstract}

Generalized Class Discovery (GCD) aims to dynamically assign labels to unlabelled data partially based on knowledge learned from labelled data, where the unlabelled data may come from known or novel classes. The prevailing approach generally involves clustering across all data and learning conceptions by prototypical contrastive learning. However, existing methods largely hinge on the performance of clustering algorithms and are thus subject to their inherent limitations. Firstly, the estimated cluster number is often smaller than the ground truth, making the existing methods suffer from the lack of prototypes for comprehensive conception learning. To address this issue, we propose an adaptive probing mechanism that introduces learnable potential prototypes to expand cluster prototypes (centers). As there is no ground truth for the potential prototype, we develop a self-supervised prototype learning framework to optimize the potential prototype in an end-to-end fashion. Secondly, clustering is computationally intensive, and the conventional strategy of clustering both labelled and unlabelled instances exacerbates this issue. To counteract this inefficiency, we opt to cluster only the unlabelled instances and subsequently expand the cluster prototypes with our introduced potential prototypes to fast explore novel classes. 
Despite the simplicity of our proposed method, extensive empirical analysis on a wide range of datasets confirms that our method consistently delivers state-of-the-art results. Specifically, our method surpasses the nearest competitor by a significant margin of \textbf{9.7}$\%$ within the Stanford Cars dataset and \textbf{12$\times$} clustering efficiency within the Herbarium 19 dataset.  We will make the code and checkpoints publicly available at \url{https://github.com/xjtuYW/PNP.git}.
\end{abstract}

\section{Introduction}
\label{sec:intro}

\label{sec:intro}
In recent years, deep learning methods ~\cite{ResNet,dosovitskiy2020image,caron2021emerging} have achieved significant success in the domain of image recognition. However, a significant challenge in achieving this success is the costly and labor-intensive process of annotating vast datasets. To mitigate these issues, recent studies have focused on leveraging unlabelled data through approaches, such as Semi-Supervised Learning (SSL)~\cite{learning2006semi,assran2021semi,berthelot2019mixmatch} and Novel Category Discovery (NCD)~\cite{han2019learning,han2021autonovel,fini2021unified}, aiming to train models effectively with less labelled data. While SSL and NCD have marked significant advancements, inherent assumptions limit their real-world applications: SSL requires unlabelled data to share the labelled data's label space, and NCD assumes all unlabelled data represent entirely new classes. To overcome these challenges and expand the applicability of NCD, Generalized Category Discovery (GCD)~\cite{vaze2022generalized} has been introduced, offering a more flexible and applicable approach in diverse settings

\begin{figure}[t]
    \centering
    \includegraphics[width=1.0\columnwidth]{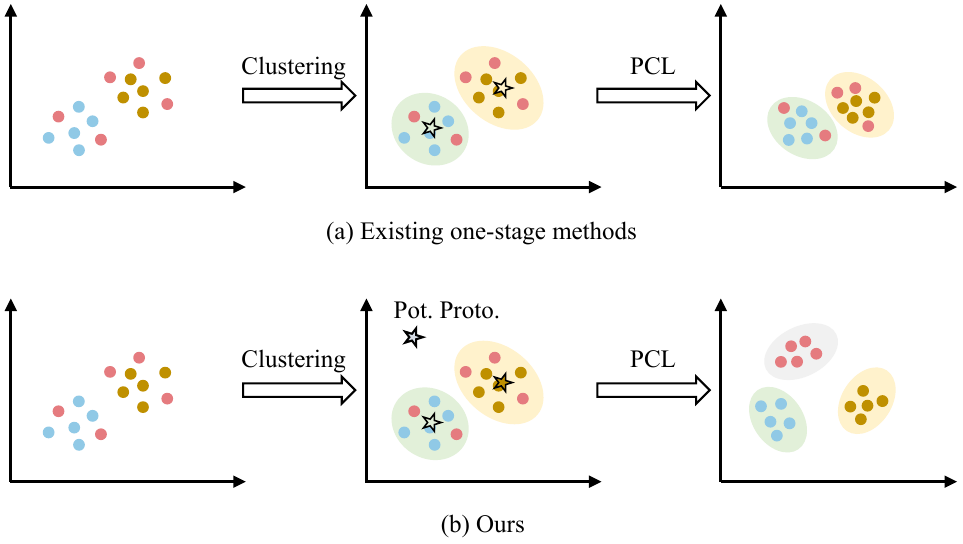}
    \caption{This illustration delineates the distinctions among various methods, where each circle represents an instance, diverse colors symbolize different classes or groups, and the pentagram means the cluster prototype (center). In contrast to (a) existing one-stage methods~\cite{pu2023dynamic,zhao2023learning}, a salient distinction of our proposed method is (b) the incorporation of potential prototypes (Pot. Proto.) to effectively identify a broader spectrum of novel classes.}
    \label{fig:motiv}
\end{figure}

GCD introduces a challenging open-world recognition task, aiming to develop a robust classification model. This model classifies unlabelled data by leveraging insights from labelled data, regardless of whether these data belong to known or new classes, without prior knowledge of the total number of categories. The principal challenges within GCD include estimating the number of categories and learning conception-level representations.

To address this task, the prevalent method in one-stage GCD works~\cite{zhao2023learning,pu2023dynamic,yang2023generalized} begins with clustering, followed by representation learning, predominantly through prototypical contrastive learning. This approach aims to align each instance's representation closely with its assigned category prototype (or cluster center). The effectiveness of this learning framework depends significantly on the clustering algorithm's ability to reveal novel categories (prototypes) and enable the learning of discriminative features. However, contemporary approaches are simultaneously limited by the inherent limitations of clustering methodologies. Notably, clusters formed by these algorithms often underestimate the actual number of categories in the early learning stage~\cite{pu2023dynamic}. The scarcity of cluster candidates inevitably results in an insufficient number of cluster prototypes, which in turn leads to blurred semantics for the clusters. Consider a dataset comprising three unique conceptions, however, the clustering algorithm's limitations can lead to the detection of only two conceptions. As illustrated in Figure~\ref{fig:motiv}(a), this outcome suggests that a minimum of one cluster will include instances from different categories, inevitably confounding the model to accurately capture the novel classes. 

To tackle this challenge, our study introduces an adaptive probing mechanism aimed at uncovering potential novel categories missed by the clustering procedure. Specifically, we achieve this by integrating learnable potential prototypes to expand existing cluster prototypes (Figure \ref{fig:motiv}(b)). These introduced prototypes serve as potential candidates to compensate for the conceptions overlooked by the clustering procedure. However, different from traditional clustering-derived prototypes, the introduced potential prototypes initially have no members, and, importantly, lack ground truth for end-to-end optimization. To address this challenge, we devise a self-distilled prototype learning strategy for refining potential prototypes. To achieve this, we initially design an asynchronous model to serve as an auxiliary encoder. Subsequently, we augment the unlabelled samples and input them into both encoders to obtain feature representations, and estimate their membership to the expanded prototypes. Finally, we encourage the predictions from respective encoders to approach each other to effectively optimize the potential prototypes.

Moreover, the inefficiency of clustering presents a persistent challenge for one-stage GCD. Traditionally, clustering encompasses both labelled and unlabelled samples to uncover conceptions within unexplored data, significantly increasing computational demands. Our proposed prototype probing strategy, demonstrating considerable effectiveness, allows for clustering solely among unlabelled data. This approach significantly enhances efficiency. To offset the absence of labelled data in guiding the conception learning of unlabelled data, we employ a direct prototype learning approach on the labelled data, utilizing randomly initialized and learnable prototypes. The proposed method, simple yet effective, works in tandem with adaptive prototype learning, providing an effctive and efficient solution for GCD.

Our contributions are summarized as follows:
\begin{itemize}
    \item We introduce an adaptive probing mechanism that leverages learnable potential prototypes in conjunction with self-distillation to uncover potential novel categories.
    \item We propose an efficient clustering strategy that focuses exclusively on unlabelled data to optimize the computational resources.
    \item Through extensive experimentation across a wide range of datasets, we validate the effectiveness of our method and elucidate its operational mechanics.
\end{itemize}

\section{Related Work}\label{sec:related}

\begin{figure*}[t]
    \centering
    \includegraphics[width=1.99\columnwidth]{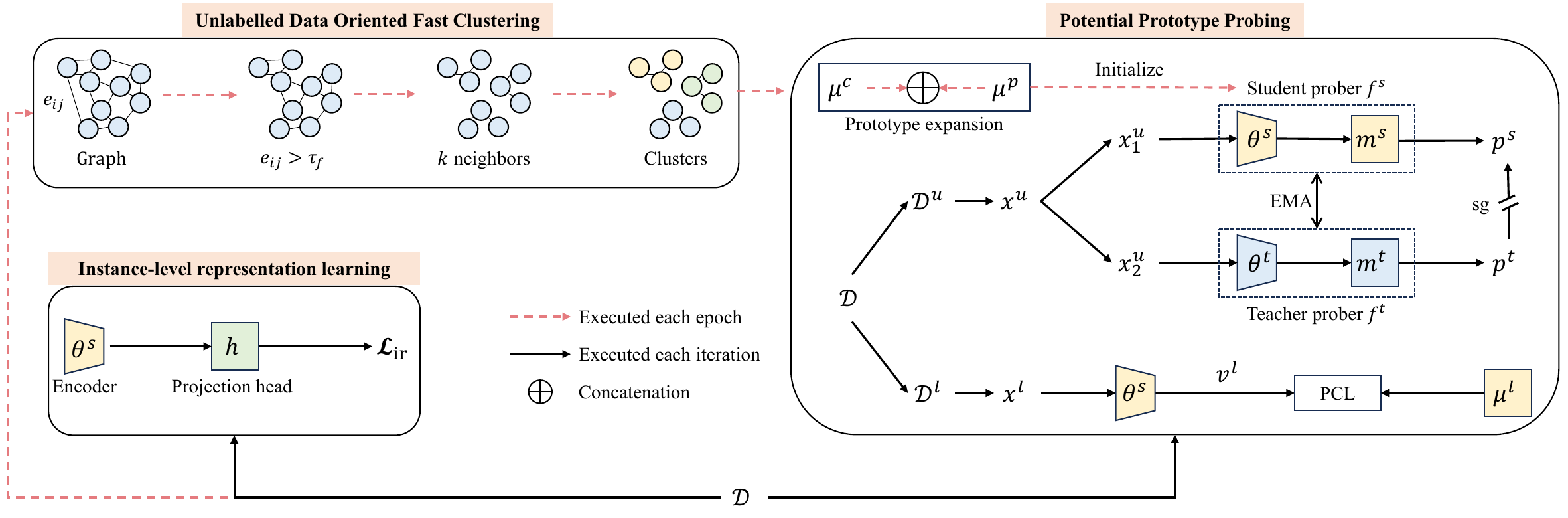}
    \vspace{-0.3cm}
    \caption{Illustration of PNP framework. The PNP framework adeptly acquires conception-level representations via potential prototype probing and further refines these representations through self-distillation learning. Initially, the potential prototype probing learns conception-level representation with $\mathcal{D}^{u}$ by prototype expansion and self-distilled prototype representation learning. Then, it learns conception-level representation with $\mathcal{D}^{l}$ by setting trainable prototypes and prototypical contrastive learning. After the training process, the framework retains solely the encoder of the student prober.} 
    \label{fig:method}
\end{figure*}

\subsection{Category Discovery}
Novel Category Discovery (NCD), a task aiming to discover new categories with the knowledge learned from labelled categories, is first formalized in DTC~\cite{han2019learning}. Rankstats~\cite{han2021autonovel} indicates that self-supervised and supervised learning help solve the NCD task. DualRank~\cite{zhao2021novel} improves Rankstats by introducing two branches to learn local and global features.
NCL~\cite{zhong2021neighborhood} and WTA~\cite{jia2021joint} employ contrastive learning to learn discriminative representations. UNO~\cite{fini2021unified} proposes a unified learning framework that combines pseudo-labeling and swapping strategies to train the model on  labelled and unlabelled data jointly. Previous transfer learning methods(\eg, KCL~\cite{hsu2018learning} and MCL~\cite{hsu2019multi}) can be also adapted to solve the NCD task. Recently, Generalized Class Discovery (GCD)~\cite{vaze2022generalized} extents NCD to a more challenging open-world setting. SimGCD~\cite{wen2023parametric} learns a parametric classification model for GCD by self-distillation. $\mu$GCD~\cite{vaze2023no} improves SimGCD by introducing strong augmentation operations and EMA strategy. PromptCAL~\cite{zhang2023promptcal} and SPTNet~\cite{wang2024sptnet} introduce auxiliary prompts to perform contrastive learning. InfoSieve~\cite{rastegar2023learn} learns a binary tree by contrastive learning to represent the category hierarchy of unlabelled data. The above GCD methods generally involve a decoupled clustering stage that utilizes the semi-supervised $k$-means (SSK) to estimate the category number. Recent work CiPR~\cite{hao2024cipr} improves the efficiency of SSK by employing selective neighbor clustering. In contrast to these methods, 
GPC~\cite{zhao2023learning}, Yang \etal~\cite{yang2023generalized}, and DCCL~\cite{pu2023dynamic} demonstrate that category number estimation and representation learning mutually benefit each other. The common framework in these one-stage methods is data clustering followed by conception learning. For example, DCCL~\cite{pu2023dynamic} alternatively employs Infomap~\cite{rosvall2008maps} to cluster all data and prototypical contrastive learning to learn representations. ORCA~\cite{cao2021openworld} and Girish \etal ~\cite{girish2021towards} propose a setting similar to GCD, but ORCA mainly focuses on the task dubbed open-world semi-supervised learning while Girish \etal ~\cite{girish2021towards} focus on the GAN task. Recently, GCD is also extended to other scenarios, such as text classification~\cite{an2024transfer,an2023generalized}, active learning~\cite{ma2024active}, and incremental learning~\cite{wu2023metagcd,zhao2023incremental,kim2023proxy}.

Previous one-stage GCD methods often suffer from insufficient prototypes for conception-level representation learning and inefficient clustering algorithms, our proposed PNP mainly focuses on solving these issues in this paper.

\subsection{Contrastive Learning}
Contrastive learning~\cite{chen2020simple,he2020momentum,li2020prototypical}, a technique that encourages representations of different augmentation views of the same instance to be similar, has been shown very effective for representation learning in self-, semi- and unsupervised learning. Recent contrastive learning methods can be roughly divided into two groups, instance-level~\cite{chen2020simple,he2020momentum,grill2020bootstrap,assran2021semi} and conception-level~\cite{caron2020unsupervised,caron2021emerging,li2020prototypical,huang2021deep,assran2022masked} contrastive learning methods. The instance-level contrastive learning methods learn the representation by comparing the similarities between the representations of positive and negative instances, while the conception-level contrastive learning methods learn the representation by comparing the similarities between the representations of instances and the prototypes, where the prototypes are either obtained by clustering~\cite{li2020prototypical,huang2021deep}  or initialized with learnable parameters~\cite{caron2020unsupervised,caron2021emerging,assran2022masked}. Among conception-level contrastive learning methods, self-distillation~\cite{caron2021emerging,assran2022masked} which minimizes the prediction distributions of two asynchronously updated models has been shown very effective for representation learning.

Inspired by these works, our study utilizes cluster prototypes to learn the conception of classes estimated by the clustering algorithm and learnable prototypes to learn the conception of unknown or potential novel classes. Further, our study employs self-distillation and instance-level representation learning to enhance conception-level representation learning.

\section{Preliminaries}\label{sec:problem}

{\noindent{\textbf{GCD Formulation.}}} Generalized category discovery (GCD) aims to adaptively assign labels to unlabelled data partially with knowledge from labelled data, where the unlabelled data may come from known or novel classes and the category number of unlabelled data is unknown. Formally, given a dataset $\mathcal{D}=\{\mathcal{D}^{l}, \mathcal{D}^{u}\}$, where $\mathcal{D}^{l}=\{(x_i^l, y_i^l)\}_{i=1}^{N}\in{\chi\times\mathcal{Y}^l}$ denotes the labelled data and $\mathcal{D}^{u}=\{(x_i^u, y_i^u)\}_{i=1}^{M}\in{\chi\times\mathcal{Y}^u}$ denotes the unlabelled data. The label space $\mathcal{Y}^l$ of $\mathcal{D}^{l}$ and the label space $\mathcal{Y}^u$ of $\mathcal{D}^{u}$ satisfy $\mathcal{Y}^l\subset\mathcal{Y}^u$, but instances of $\mathcal{D}^{u}$ are not observed in $\mathcal{D}^{l}$. During the training phase, $\mathcal{D}^{l}$ and $\mathcal{D}^{u}$ are used to train the model. The goal of GCD is to cluster instances with the same category in $\mathcal{D}^{u}$ into the same group and assign the label. The key challenges in GCD are category number estimation and representation learning.

{\noindent{\textbf{One-Stage GCD Revisit.}}}  To solve this task, existing one-stage GCD works~\cite{pu2023dynamic,zhao2023learning} usually perform clustering on both $\mathcal{D}^{u}$ and $\mathcal{D}^{l}$ to estimate the category number and get the cluster centers (prototypes) in each training epoch. Then, they utilize the clustering results to learn the conception-level representations using prototypical contrastive learning. Concretely, let $v$ denotes the encoded feature and $\mu=\{{\mu}_1,...,{\mu}_{K^e}\}$ denotes the prototypes obtained from clustering results of $\{\mathcal{D}^{u},\mathcal{D}^{l}\}$.
The primary learning objective is given by:
\begin{equation}
\label{eq:pcl}
     \mathcal{L}_\text{cr} = \frac{1}{|B|}\sum_{q\in{\mathcal{Q}(i)}}-\text{log}\frac{\exp({v_q}\cdot{\mu}_{i}/\tau)}{\sum_{j=1}^{K^e} \exp(v_q \cdot {\mu}_{j}/\tau)},
\end{equation}
where $\mathcal{Q}(i)$ denotes instances with the label $i$, $\tau$ is the temperature used to adjust the magnitude of the prediction distribution, and $\cdot$ refers to dot production. 
\section{Methodology}\label{sec:method}

In response to the weakness of existing one-stage GCD models, \ie, insufficient prototypes for conception-level representation learning and inefficient clustering algorithm, we propose a novel framework of Probing New Prototype (PNP). Particularly, a potential prototype probing mechanism is proposed to expand the shrunk prototypes caused by the clustering algorithm, and an effective strategy of solely clustering on the unlabelled data for a fast class discovery.

Figure~\ref{fig:method} illustrates the detailed architecture, PNP includes three components, a student prober $f^s$ consisting of an encoder parameterized by ${\theta}^s$ and a memory buffer ${m}^s$ with fixed size $K^t$, a teacher prober $f^t$ that shares the same structure as $f^s$, and a projection head $h$. The overall network is organized as two branches with the encoder of $f^s$ shared: instance representation learning and potential prototype probing, where the instance representation learning branch, formed by the encoder of $f^s$ and projection head $h$, is in charge of learning the discriminative features of samples. The potential prototype probing ($\text{p}^3$) utilizes three components to align the samples to the category (known or unknown) prototypes, which is a key module for class discovery. p$^3$ branch first conducts a fast clustering on the unlabelled samples, the subsequent p$^3$ module takes the clustering prototype and expands it with potential prototype candidates. Finally, the p$^3$ branch is optimized by the objectives from the devised self-distillation learning. Subsequently, we introduce the component of p$^3$ branch, which is the key module of this study. 

\subsection{Unlabelled Data Oriented Fast Clustering}
\label{subsec:fast_c}
For efficiency, PNP utilizes $\mathcal{D}^{u}$ instead of $\{\mathcal{D}^{l}, \mathcal{D}^{u}\}$ to accelerate the clustering process. Concretely, we adopt the Infomap~\cite{rosvall2008maps} as the clustering method in this paper. Infomap is an efficient and effective unsupervised clustering algorithm, which is quite in line with our target. Let $v^u=f^s(x^u;{\theta}^s)$ denote the encoded feature of $\mathcal{D}^{u}$. In each training epoch, PNP first constructs a graph $\mathcal{G}=\{\mathcal{I},\mathcal{E}\}$, where $\mathcal{I}$ refers to unlabelled instances and $\mathcal{E}$ refers to the union of edges weighted by the cosine similarity between the corresponding instances. 
Then, PNP filters the edges by:
\begin{equation}
    e_{ij}=\left\{
            \begin{array}{ll}
                e_{ij}, & \text{if} \: e_{ij} \textgreater {\tau}_f \:\text{and}\: i\neq j, \\
                 0, &  \text{others,}
            \end{array}
\right.
\end{equation}
where $e_{ij}=v_i^u \cdot v_j^u$ refers to the edge weight between instance $i$ and $j$, ${\tau}_f$ is a hyper-parameter used to control relation reliability between two different instances.
Next, PNP employs $k$NN to keep top-$k$ edges for each instance to accelerate the clustering process further.
In the end, with the filtered graph, the category number estimation $K^e$ and clustering results are obtained with Infomap.

\subsection{Potential Prototype Probing.}
\label{subsec:ppp}
With the clustering results, PNP first gets the prototype (mean feature) of each cluster followed by prototype expansion. Then, PNP conducts self-distilled prototype representation learning to learn conception-level representations.

\noindent\textbf{Prototype Expansion with Potential Candidates.} Let $\mathcal{Y}^e$ denote the label space of clustered data, the cluster prototypes ${\mu}^c=\{{\mu}_0,...,{\mu}_{K^e}\}$ can be obtained by
\begin{equation}
    \mu_i=\frac{1}{|\mathcal{Y}^e_i|}\sum_{q\in{\mathcal{Q}(i)}}\frac{v^u_q}{\Vert v^u_q \Vert_2},
\end{equation}
where $\mathcal{Y}^e_i$ denotes the label space of the $i$-th cluster. Then, PNP initializes several trainable tensors ${\mu}^p\in\mathbb{R}^{(K^t-K^e)\times d}$ as potential prototypical candidates and concatenates ${\mu}^p$ with 
the ${\mu}^c$, where $d$ is the dimension of the encoded feature. 

\noindent\textbf{Self-distilled Prototype Representation Learning}. Since there is no ground truth available for the potential prototype learning, we adopt a self-distilled manner to optimize the potential prototypes. Formally, let $x^u_1$ and $x^u_2$ denote two augmentation views of same instance $x^u$ in $\mathcal{D}^{u}$, we next initialize the memory buffer ${m}^s$ of student prober $f^s$ with the concatenation of ${\mu}^p$ and ${\mu}^c$. Meanwhile, the memory buffer $m^t$ of teacher prober $f^t$ is also initialized as the concatenation of${\mu}^p$ and ${\mu}^c$. Then, PNP feeds $x^u_1$ into $f^s$ to get the prediction $p^s$ by:
\begin{equation}
\label{eq:stu_pred}
    p^s = \text{softmax}(\frac{v^u_1 \cdot m^s}{{\tau}}),
\end{equation}
where $v^u_1=f^s(x^u_1;{\theta}^s)$.
Meanwhile, PNP feeds $x^u_2$ into $f^t$ to obtain the soft prediction $p^t$ by:
\begin{equation}
    p^{t} = \text{softmax}(\frac{v^u_2 \cdot m^t}{{\tau}_{t}}),
\end{equation}
where $v^u_2=f^t(x^u_2;{\theta}^t)$ and ${\tau}_{t}$ denotes the temperature used to sharp the prediction distribution.
After obtaining $p^t$ and $p^s$, the conception-level representation learning objective with unlabelled data is given by
\begin{equation}
    \mathcal{L}_\text{cru}= \frac{1}{|B|}\sum{p^{t}\text{log}(p^s)}+\gamma{R(\overline{p})},
\end{equation}
where $R(\overline{p})=\overline{p}\text{log}(\overline{p})$, $\overline{p}=\frac{1}{|B|}p^s$ refers to the mean-entropy maximization regularizer~\cite{assran2021semi} and $\gamma$ is the hyperparameter used to control the contribution of the regularization loss. 

To stabilize the learning, we stop the gradient (sg) of $p^{t}$ and update the parameters ${\Theta}_t=\{{\theta}^t,m^t\}$ of the teacher prober $f^t$ in an Exponential Moving Average (EMA) manner~\cite{grill2020bootstrap}, \ie, ${\Theta}_t=\omega(t){\Theta}_t+(1-\omega(t)){\Theta}$, where ${\Theta}$ refers to the trainable parameters of the student prober $f^s$, and the hyperparameter $\omega(t)=\omega_\text{max}-(1-\omega_\text{min})\text{cos}(\frac{\pi{t}}{T}+1)/2$ is updated in cosine decay manner~\cite{vaze2023no}.

\noindent\textbf{Conception Modeling with Known Data.} 
Labeled data offers definitive conceptual cues, playing a pivotal role in guiding the overall network learning. To integrate the reliable knowledge,
PNP treats instances with the same labels in $\mathcal{D}^{l}$ as a cluster and employs prototypical contrastive learning to implicitly guide the conception learning of $\mathcal{D}^{u}$. 
Concretely, let ${\mu}^l\in{|\mathcal{Y}^l|\times d}$ denotes the trainable prototypes. 
For each instance $x^l$ in $\mathcal{D}^{l}$, PNP first feeds $x^l$ into the encoder of $f^s$ and then utilizes Eq.\ref{eq:pcl} and ${\mu}^l$ to calculate another conception-level representation learning objective $\mathcal{L}_\text{crl}$.

Finally, the overall conception-level representation learning objective $\mathcal{L}_\text{cr}$ is given by
\begin{equation}
    \mathcal{L}_\text{cr} = {\alpha}_1 \mathcal{L}_\text{cru} + (1-{\alpha}_1)\mathcal{L}_\text{crl},
\end{equation}
where ${\alpha}_1$ is the hyperparameter used balance the contribution of $\mathcal{L}_\text{cru}$ and $\mathcal{L}_\text{crl}$.

\subsection{Training and Inference.}
\noindent\textbf{Network Training.} To learn the discriminative instance representation, PNP adopts the instance-level learning objectives, comprising the supervised term for known samples and the unsupervised part for unknown samples:
\begin{equation}
    \mathcal{L}_\text{ir}={\beta}_1 \mathcal{L}_\text{sup} + (1-{\beta}_1) \mathcal{L}_\text{unsup},
\end{equation}
where ${\beta}_1$ refers to the hyperparameter used to balance the contribution of supervised contrastive learning loss $\mathcal{L}_\text{sup}$ and unsupervised contrastive learning loss $\mathcal{L}_\text{unsup}$. More concretely, let $x^1$ and $x^2$ denote the two augmentation views of the same instance in a mini-batch $B$, $z^1$ and $z^2$ denote the projected features given by passing $x^1$ and $x^2$ to the encoder of student prober $f^s$ and the projection head $h$, respectively. The  $\mathcal{L}_\text{sup}$ is given by
\begin{equation}
    \mathcal{L}_\text{sup}=\frac{1}{|B^l|}\sum_{i}\frac{1}{|\mathcal{N}(i)|}\sum_{q\in{\mathcal{N}(i)}}-\text{log}\frac{{\exp(z^1_i \cdot z^1_q/{\tau}_{{r}})}}{\sum_{j\neq{i}}{\exp(z^1_i \cdot z^1_j/{\tau}_{{r}})}},
\end{equation}
where $\mathcal{N}(i)$ denotes the other instances which have the same label as instance $i$ and ${\tau}_r$ refers to the temperature.
The $\mathcal{L}_\text{unsup}$ is analogously calculated:
\begin{equation}
    \mathcal{L}_\text{unsup}=\frac{1}{|B|}\sum-\text{log}\frac{{\exp(z^1_i \cdot z^2_i/{\tau}_{{r}})}}{\sum_{j\neq{i}}{\exp(z^1_i \cdot z^1_j/{\tau}_{{r}})}}.
\end{equation}

In summary, the full objectives for network training reads:
\begin{equation}
    {\Theta}^* = \mathop{\arg\min}\limits_{{\Theta}}\mathcal{L}_{\text{cr}} + \mathcal{L}_{\text{ir}}.
\end{equation}

\noindent\textbf{Inference.} When the training is done, PNP only keeps the encoder of student prober $f^s$. In the inference stage, there are three steps to assign labels for $\mathcal{D}^{u}$: 1) the representation of each instance in $\mathcal{D}^{u}$ is first extracted using the encoder of student prober $f^s$, 2) using fast clustering algorithm in sec.~\ref{subsec:fast_c} to cluster the encoded features, 3) assigning the labels by matching predicted cluster assignments to the actual class labels.

\section{Experiments}\label{sec:exp}

\begin{table*}[t]\small
    \centering
    \begin{tabular}{cccccccc}
    \toprule
        {} & CIFAR10 & CIFAR100 & ImageNet100 & Herbarium 19 & CUB200 & Stanford Cars & FGVC-Aircraft \\
    \midrule
        {$|\mathcal{Y}^{l}|$} & 5   & 80  & 50  & 341 & 100 & 98 & 50 \\
        {$|\mathcal{Y}^{u}|$} & 10  & 100  & 100 & 683 & 200 & 196 & 100 \\
    \midrule
        {$|\mathcal{D}^{l}|$} & 12.5k   & 20k & 31.9k  & 8.9k & 1.5k & 2.0k & 1.7k \\
        {$|\mathcal{D}^{u}|$} & 37.5k   & 30k & 95.3k & 25.4k & 4.5k & 6.1k & 5.0k \\
    \bottomrule
    \end{tabular}
    \caption{Statistics of the datasets used in our experiments.}
    \label{tab:data_info}
\end{table*}

\begin{table*}[ht]\small
    \centering
    \begin{tabular}{c|c|c|ccc|ccc|ccc}
    \toprule
    \multirow{2}{*}{Methods} & \multirow{2}{*}{Ref.} & \multirow{2}{*}{Known $K^u$} & \multicolumn{3}{c|}{CUB200} & \multicolumn{3}{c|}{Stanford Cars} & \multicolumn{3}{c}{FGVC-Aircraft} \\
    \cmidrule(r){4-6}  \cmidrule(r){7-9} \cmidrule(r){10-12}
    {} & {} & {} & {All} & {Old} & {New}  & {All} & {Old} & {New} & {All} & {Old} & {New} \\
    \midrule
         GCD~\cite{vaze2022generalized}         & {(CVPR'22)} & \ding{51}  & 51.3 & 56.6 & 48.7 & 39.0 & 57.6 & 29.9 & 45.0 & 41.1 & 46.9\\
         CiPR~\cite{hao2024cipr}                & {(TMLR'24)} & \ding{51}  & 57.1 & 58.7 & 55.6 & 47.0 & 61.5 & 40.1 & {-}  & {-} & {-} \\ 
         SimGCD~\cite{wen2023parametric}        & {(ICCV'23)} & \ding{51}  & 60.3 & 65.6 & 57.7 & 53.8 & 71.9 & 45.0 & 54.2 & 59.1 & 51.8 \\
         PromptCAL~\cite{zhang2023promptcal}    & {(ICCV'23)} & \ding{51}  & 62.9 & 64.4 & 62.1 & 50.2 & 70.1 & 40.6 & 52.2 & 52.2 & 52.3 \\
         $\mu$GCD~\cite{vaze2023no}             & {(NeurIPS'23)} & \ding{51}  & 65.7 & 68.0 & {64.6} & \underline{56.5} & 68.1 & \textbf{50.9} & 53.8 & 55.4 & \underline{53.0} \\
         SPTNet~\cite{wang2024sptnet}           & {(ICLR'24)} & \ding{51}  & \underline{65.8} & \underline{68.8} & \underline{65.1} & \textbf{59.0} & \textbf{79.2} & \underline{49.3} & \textbf{59.3} & \underline{61.8 }& \textbf{58.1} \\
         InfoSieve~\cite{rastegar2023learn}     & {(NeurIPS'23)} & \ding{51}  & \textbf{69.4} & \textbf{77.9} & \textbf{65.2} & {55.7} & \underline{74.8} & 46.4 & \underline{56.3} & \textbf{63.7} & 52.5 \\
         \midrule
         GCD$^*$~\cite{vaze2022generalized}     & {(NeurIPS'23)} & \ding{55}   & 47.1 & 55.1 & 44.8 & 35.0 & 56.0 & 24.8 & 40.1 & 40.8 & 42.8 \\
         GPC~\cite{zhao2023learning}            & {(ICCV'23)} & \ding{55}   & 52.0 & 55.5 & 47.5 & 38.2 & 58.9 & 27.4 & \underline{43.3} & 40.7 & \underline{44.8} \\
         Yang \etal~\cite{yang2023generalized}  & {(NeurIPS'23)} & \ding{55}& 58.0 & \underline{65.0} & 43.9 & \underline{47.6} & \textbf{70.6} & 33.8 & {-}  & {-} & {-} \\ 
         DCCL$^*$~\cite{pu2023dynamic}          & {(ICCV'23)} & \ding{55}   & \underline{63.5} & {60.8} & \underline{64.9} & 43.1 & 55.7 & \underline{36.2} & 42.7 & \underline{41.5} & 43.4 \\
         
    
         PNP        & {Ours} & \ding{55} & \textbf{67.4} & \textbf{69.2} & \textbf{66.5} & \textbf{57.3} & \underline{70.2} & \textbf{51.1} & \textbf{47.0} & \textbf{49.5} & \textbf{45.7} \\
         
   \bottomrule  
    \end{tabular}
    \caption{Results on the Semantic Shift Benchmark, where $^*$ denotes our reproduced results or from their publicly available results, $-$ denotes no results reported in the corresponding works, and bold and \underline{underlined} values represent the best and \underline{second} best results with/without taking the category number $K^u$ as the prior, respectively. }
    \label{tab:ssb}
\end{table*}


\begin{table*}[ht]\footnotesize
    \centering
    \begin{tabular}{c|c|c|ccc|ccc|ccc|ccc}
    \toprule
    \multirow{2}{*}{Methods} & \multirow{2}{*}{Ref.} & \multirow{2}{*}{Known $K^u$} & \multicolumn{3}{c|}{CIFAR10} & \multicolumn{3}{c|}{CIFAR100} & \multicolumn{3}{c|}{ImageNet100} & \multicolumn{3}{c}{Herbarium 19} \\
    \cmidrule(r){4-6}  \cmidrule(r){7-9} \cmidrule(r){10-12} \cmidrule(r){13-15}
    {} & {} & {} & {All} & {Old} & {New}  & {All} & {Old} & {New} & {All} & {Old} & {New} & {All} & {Old} & {New} \\
    \midrule
         GCD~\cite{vaze2022generalized}  & {(CVPR'22)}   & \ding{51}  & 91.5 & \textbf{97.9} & 88.2 & 73.0 & 76.2 & 66.5 & {72.7}  & {91.8} & {63.8} & 35.4 & 51.0 & 27.0 \\
         SimGCD~\cite{wen2023parametric} & {(ICCV'23)}   & \ding{51}  & 97.1 & 95.1 & {98.1} & 80.1 & 81.2 & \underline{77.8} & {83.0} & {93.1} & {77.9} & \textbf{44.0} & \underline{58.0} & \textbf{36.4} \\
         InfoSieve~\cite{rastegar2023learn} & {(NeurIPS'23)} & \ding{51}  & 94.8 & 97.7 & 93.4  & 78.3 & 82.2 & 70.5 & {80.5} & \textbf{93.8} & {73.8} & 41.0 & {55.4} & 33.2 \\
         CiPR~\cite{hao2024cipr}            & {(TMLR'24)} & \ding{51}  & \underline{97.7} & \underline{97.5} & 97.7 & \textbf{81.5} & 82.4 & \textbf{79.7} & 80.5 & 84.9 & \underline{78.3} & 36.8 & 45.4 & 32.6\\
         SPTNet~\cite{wang2024sptnet} & {(ICLR'24)} & \ding{51}  & 97.3 & 95.0 & \textbf{98.6} & \underline{81.3} & \textbf{84.3} & 75.6 & \textbf{85.4} & \underline{93.2} & \textbf{81.4} & \underline{43.4} & \textbf{58.7} & \underline{35.2} \\
         PromptCAL~\cite{zhang2023promptcal} & {(ICCV'23)} & \ding{51}  & \textbf{97.9} & {96.6} & \underline{98.5} & {81.2} & \underline{84.2} & 75.3 & \underline{83.1} & {92.7} & \underline{78.3} & - & - & - \\

         \midrule
         GCD$^*$~\cite{vaze2022generalized} & {(CVPR'22)} & \ding{55}   & 88.6 & 96.2 & 84.9 & 73.0 & 76.2 & 66.5 & {72.7} & {91.8} & {63.8} & 30.9 & 36.3 & 28.0 \\
         GPC~\cite{zhao2023learning}       & {(ICCV'23)} & \ding{55}   & 90.6 & \textbf{97.6} & 87.0 & 75.4 & \textbf{84.6} & 60.1 & {75.3} & \textbf{93.4} & {66.7} & 36.5 & 51.7 & 27.9 \\
         Yang \etal~\cite{yang2023generalized} & {(NeurIPS'23)} & \ding{55}& \underline{92.3} & 91.4 & \underline{94.4} & \underline{78.5} & 81.4 & \underline{75.6} & \underline{81.1} & 80.3 & \textbf{81.8} & 36.3 & \underline{53.1} & 30.7 \\
         DCCL$^*$~\cite{pu2023dynamic}     & {(ICCV'23)} & \ding{55}   & \textbf{96.3} & \underline{96.5} & \textbf{96.9} & 75.3 & 76.8 & 70.2 & {80.5} & {90.5} & {76.2} & \underline{39.8} & 50.5 & \textbf{34.1} \\
         PNP                        & {Ours} & \ding{55}   & {90.2} & {95.0} & {87.8} & \textbf{80.0} & \underline{81.7} & \textbf{76.3} & \textbf{82.4} & \underline{92.9} & \underline{77.2} & \textbf{41.1} & \textbf{56.2} & \underline{32.9} \\
         
   \bottomrule  
    \end{tabular}
    \caption{Results on generic image recognition datasets and more challenging fine-grained and long-tailed datasets, where $^*$ denotes our reproduced results or from their publicly available results, and bold and \underline{underlined} values represent the best and \underline{second} best results with/without taking the category number $K^u$ as the prior, respectively. }
    \label{tab:cch}
\end{table*}

\begin{table*}\begin{center}\footnotesize
    \begin{tabular}{ccccccc}
    \toprule 
    Method & CUB200 &  Stanford Cars & FGVC-Aircraft & CIFAR10 & CIFAR100 & Herbarium 19 \\
    \midrule
    DCCL~\cite{pu2023dynamic} & {53s} & {4.02m} & {3.97m}  & {15.33m} & {9.87m} & {24.78m} \\
    PNP(Ours)                 & {23s} & {42s}   & {59s}    & {7.25m}  & {2.87m} & {2.05m} \\
    
    \bottomrule
    \end{tabular}
    \caption{Statistics of time cost for clustering in our proposed method and DCCL, where the experiments are conducted using the same GPU and faiss~\cite{johnson2019billion} to accelerate the graph construction process.}
    \label{tab:time}
    \end{center}
\end{table*}

\section{Experiments}\label{sec:exp}

\noindent\textbf{Datasets.}
Seven benchmarks are employed to make a comprehensive performance evaluation, including three generic datasets CIFAR10~\cite{krizhevsky2009learning}, CIFAR100~\cite{krizhevsky2009learning}, and ImageNet100~\cite{deng2009imagenet}, three recent proposed Semantic Shift Benchmark~\cite{vaze2022openset} (SSB) CUB~\cite{cub}, Stanford Cars~\cite{krause20133d}, and FGVC-Aircraft~\cite{maji2013fine}), and the harder long-tailed and fine-grained dataset Herbarium 19~\cite{tan2019herbarium}. For each dataset, we follow~\cite{vaze2022generalized} to sample a subset of all classes as the labeled classes or ``Old'' classes, and the remaining classes as the unlabelled classes or ``New'' classes. 50$\%$ of the images of old classes constitute the labeled set $\mathcal{D}^{l}$. The remaining images of old classes and all images of new classes are used to construct the unlabelled set $\mathcal{D}^{u}$.  Table~\ref{tab:data_info} presents the detailed statistics of the datasets.

\noindent\textbf{Evaluation Protocol.}
At test time, we follow~\cite{vaze2022generalized} to measure the clustering accuracy (ACC) between the ground truth $y^*$ and the model's predictions $\hat{y}$ for each dataset, the ACC is calculated as ACC$=\frac{1}{M}\sum_{i=1}^{M}\mathbb{I}(y^*_i==m(\hat{y}_i)$, where $M=|\mathcal{D}^u|$ and $m$ refers to the optimal permutation for matching predicted cluster assignment to the actual class label.

\noindent\textbf{Implementation Details.}
Following ~\cite{vaze2022generalized,wen2023parametric,pu2023dynamic}, we adopt ViT-B/16 pretrained by DINO~\cite{caron2021emerging} as the encoder and take the encoder's output \texttt{[CLS]} token with a dimension of 768 as the feature representation. We construct the projection head using three Linear layers, where the hidden dimension is set to 2048. We set the size of the memory buffer to 4$|\mathcal{Y}^l|$, the value of $k$NN to 20 for FGVC-Aircraft, 10 for other fine-grained datasets, 250 for CIFAR100, 2000 for CIFAR10, and 1000 for ImageNet100. Following previous GCD works~\cite{vaze2022generalized,pu2023dynamic,zhao2023learning}, we set the last layer of the encoder to be trainable and freeze other layers. We train the model with a batch size of 128 for 200 epochs and the learning rate is set to 0.1, decayed with the cosine schedule~\cite{loshchilov2017sgdr}. Align with~\cite{wen2023parametric}, the loss balancing factors $\alpha_1$, $\beta_1$, and $\gamma$ are set to 0.65, 0.35, and 2, respectively. For representation learning objectives, $\tau$ is set to 0.1, $\tau_t$ is initialized using a value of 0.07 and warmed up to 0.04 using the cosine schedule in the first 30 epochs, and $\tau_r$ is set to 1. For clustering, following~\cite{pu2023dynamic}, we adopt faiss~\cite{johnson2019billion} to accelerate the similarity calculation process and set $\tau_f$ to 0.6. EMA updating schedule keep consistent with~\cite{vaze2023no}, we set $\omega_\text{min}$ to 0.7 and $\omega_\text{max}$ to 0.99, respectively. More details will be given in \url{https://github.com/xjtuYW/PNP.git}. All experiments are conducted on one NVIDIA GeForce RTX 2080 Ti.

\subsection{Comparison with State-of-the-Art.}

\noindent{\textbf{Clustering Accuracy.}} Table~\ref{tab:ssb} and Table~\ref{tab:cch} show the clustering accuracy of different methods on the SSB, CIFAR10, CIFAR100, ImageNet100, and Herbarium19, respectively. We can see that by \textit{taking the category number as a prior} which is \textit{impractical or contradictory} to the setting of the GCD task, methods, such as InfoSieve~\cite{rastegar2023learn}, achieve shining performance. In contrast to these methods, though a relatively lower performance is obtained, the one-stage methods do not take the category number as a prior. Therefore, we mainly concentrate on the performance of one-stage methods in the following. Concretely, as shown in Table~\ref{tab:ssb}, our proposed PNP almost consistently outperforms other one-stage methods on SSB. Particularly, on the Stanford Cars, PNP outperforms the second-best one-stage method proposed by Yang \etal~\cite{yang2023generalized} by a large margin of \textbf{9.7$\%$} on ''All`` classes. Further, as shown in Table~\ref{tab:cch}, except for CIFAR10, though the highest clustering accuracy on ``Old'' and ``New'' classes changes between PNP and other one-stage methods, PNP archives the best performance on the ``All'' classes.

\noindent{\textbf{Efficiency Comparison.}} Table~\ref{tab:time} shows the time cost for clustering of our proposed PNP and DCCL. Compared to the time spent on clustering in DCCL which also adopts the Infomap~\cite{rosvall2008maps}, less time is spent by PNP indicating that the clustering method proposed by PNP is more efficient than DCCL. Particularly, on Herbarium19, PNP achieves over \textbf{12$\times$} clustering efficiency compared to DCCL.

\subsection{Ablation Studies}

\begin{table}\footnotesize
    \begin{center}
    \begin{tabular}{ccc|ccc|ccc}
    \toprule
         \multirow{2}{*}{EMA} & \multirow{2}{*}{PP} & \multirow{2}{*}{TPer}  & \multicolumn{3}{c|}{CUB200} & \multicolumn{3}{c}{CIFAR100} \\
    \cmidrule{4-9}
         {}      &    {}     & {}           & {All} & {Old} & {New}  & {All} & {Old} & {New}\\
    \midrule
         {}  & {} & {}                      & {61.1}  & {65.8} & {58.7} & {74.6}  & {80.6} & {62.7} \\
         {}  & {\ding{51}} & {}             & {64.6}  & {64.4} & {64.7} & {75.5}  & {78.2} & {70.2} \\
         {}  & {} & {\ding{51}}            & {63.1}  & {65.1} & {62.1} & {76.5}  & {82.8} & {63.8} \\
         {}  & {\ding{51}} & {\ding{51}}    & {65.5}  & {65.0} & {65.7} & {78.8}  & {80.6} & {75.4} \\
         {\ding{51}}  & {} & {\ding{51}}            & {63.0}  & {65.1} & {61.7} & {78.4}  & {84.6} & {66.1} \\
         {\ding{51}}  & {\ding{51}} & {\ding{51}}   & {67.4}  & {69.2} & {66.5} & {80.0}  & {81.7} & {76.3} \\
    \bottomrule
    \end{tabular}
    \caption{Ablation studies on CUB200 and CIFAR100, where \textbf{PP} refers to the potential prototypes and \textbf{TPer} refers to the teacher prober.}
    \label{tab:ablat}
    \end{center}
\end{table}

 To validate the effectiveness of each component in PNP, we conduct six experiments on CUB200 and CIFAR100 and report the corresponding clustering accuracy in Table~\ref{tab:ablat}. Row 1 removes all components to provide a baseline. Row 2 uses only the potential prototypes (\textbf{PP}) and achieves better performance than the baseline, this indicates that using PP is helpful for conception-level representation learning. Row 3 uses only the teacher prober (\textbf{TPer}), \ie, using TPer to generate soft predictions to self-distilled prototype representation learning, better performance is achieved compared to the baseline demonstrating the effectiveness of TPer. Row 4 combines PP and TPer, the higher clustering accuracy is obtained compared to that given by using only PP (row 2) or TPer (row 3), suggesting that PP and TPer mutually benefit each other. Row 5 updates TPer using EMA but removes PP, weaker performance is obtained compared to that given by using all components (row 6), this result further validates the effectiveness of PP. Meanwhile, when combining PP and TPer, further introducing EMA (row 6) achieves better performance than that given by removing EMA (row 4) indicating that EMA is effective. In summary,  the experimental results demonstrate the effectiveness of PP, TPer, and EMA.

\subsection{Analysis}
\label{subsec:ana}

\subsubsection{Size of Memory Buffer.}

\begin{figure}[t]
    \centering
    \includegraphics[width=1.\columnwidth]{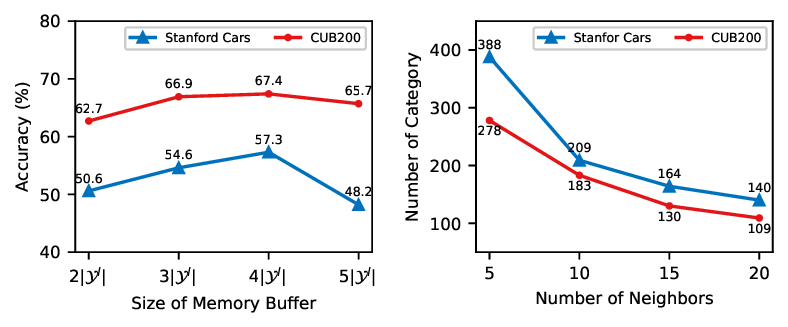}
    \caption{(left:) The overall clustering accuracy given by different sizes of memory buffer, where $|\mathcal{Y}^l|$ indicates the category number of labelled data. (right:) The overall clustering accuracy (ALL) given by different numbers of neighbors used to filter the graph in the clustering process.}
    \label{fig:hypers_pp}
\end{figure}

The left subfigure of Figure~\ref{fig:hypers_pp} shows the clustering accuracy on ``ALL'' classes given by setting different sizes of memory buffer on CUB200 and Stanford Cars. We can see that when the size of memory buffer varies from 2$|\mathcal{Y}^l|$ to 4$|\mathcal{Y}^l|$, the clustering accuracy is improved gradually. When the size of memory buffer is set to 5$|\mathcal{Y}^l|$, the performance is dropped. The results indicate that our proposed PNP prefers a relatively larger size of memory buffer. Particularly, setting the size of memory buffer to 4$|\mathcal{Y}^l|$ helps our proposed method achieve the best clustering accuracy on ``ALL'' classes.

\begin{figure*}[t]
    \centering
    \includegraphics[width=1.85\columnwidth]{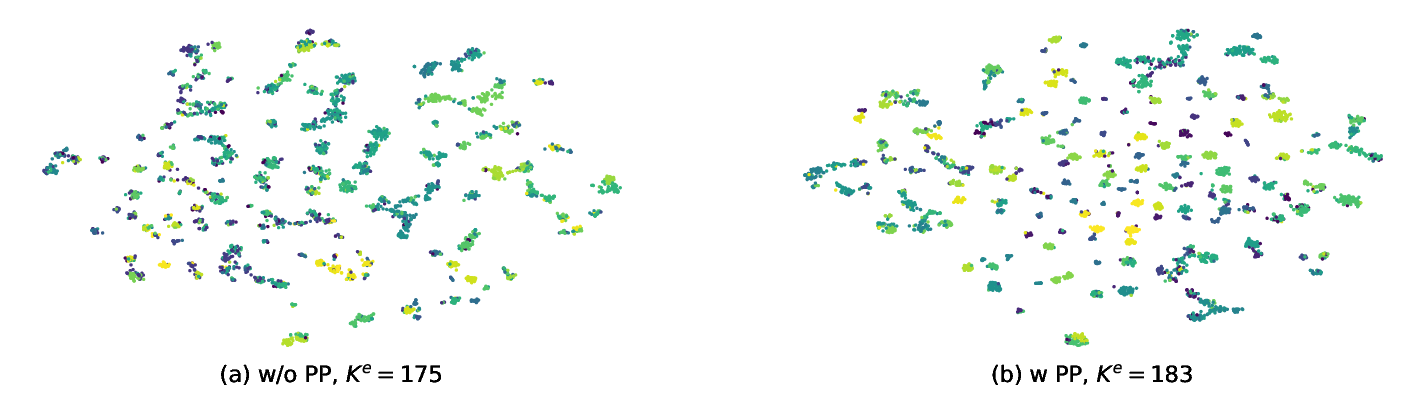}
    \caption{t-SNE visualization of instances in CUB200 for features generated by (a) removing and (b) using potential prototypes (PP), where $K^e$ refers to the estimated class or cluster number.}
    \label{fig:tsne}
\end{figure*}

\begin{figure*}[t]
    \centering
    \includegraphics[width=1.9\columnwidth]{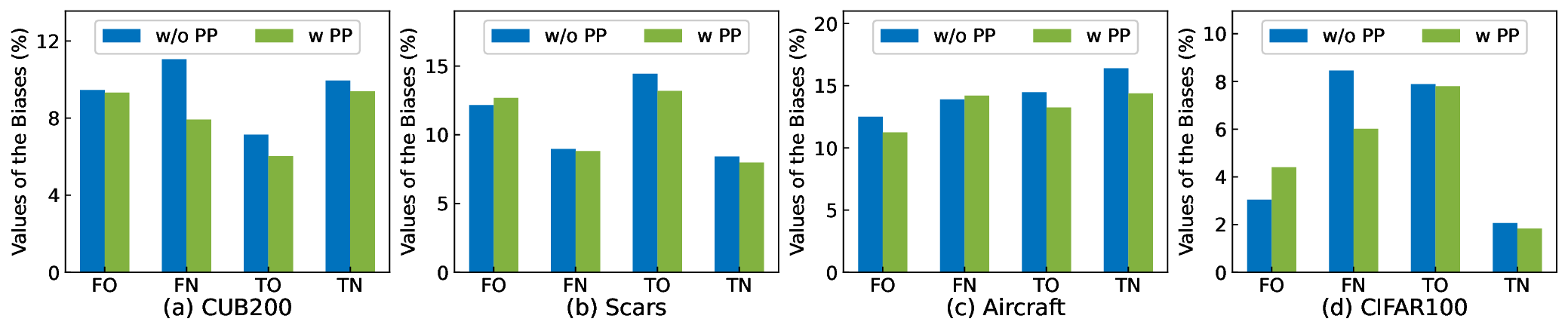}
    \caption{Prediction biases between Old/New classes with/without using potential prototypes (PP).}
    \label{fig:ablat_vp}
\end{figure*}

\subsubsection{Number of Neighbors.}

In the clustering process, PNP filters the constructed graph by setting a threshold and KNN. To explore the influence of the number of neighbors used in KNN on the category number estimation, we conduct several experiments on CUB200 and Stanford Cars and report the corresponding results in the right subfigure of Figure~\ref{fig:hypers_pp}. We can see that the estimated category number decreases as the number of neighbors increases, this indicates that retaining more edges in the graph would encourage bigger but fewer clusters. 

\subsubsection{Visualization.}
To provide an intuition of the role of potential prototypes (PP), we first collect the features generated by removing PP and using PP. Then, we employ t-SNE~\cite{van2008visualizing} to project the collected features into a 2-dimensional space and compare the two projection results. As shown in Figure~\ref{fig:tsne}, we find that removing PP results in a smaller cluster number ($K^e=175$) compared to that given by using PP ($K^e=183$), which is closer to the real categories of 200. This means that using PP helps discover more classes. Further, as shown in the center parts of Figure~\ref{fig:tsne}(a) and Figure~\ref{fig:tsne}(b), we observe that using PP leads to relatively clearer boundaries, this suggests that \textit{using PP helps reduce inter-class confusion}.

\subsubsection{Inter-Class Prediction Bias.}
To investigate how potential prototypes (PP) reduce inter-class confusion, we conduct several experiments on  CUB200, Stanford Cars, FGVC-Aircraft, and CIFAR100 to observe the inter-class prediction bias.
Align with~\cite{wen2023parametric}, the prediction biases mainly consist of four parts, False Old (FO), False New (FN), True Old (TO), and True New (TN), where FO indicates that the new category instances are predicted as old category instances, FN means that the old category instances are predicted as new category instances, TO implies that the old category instances are predicted as other old category instances, and TN represents that the new category instances are predicted as other new category instances. 
Because using PP results in a larger cluster number, we select classes that exist in both the clustering results given by removing and using PP to mitigate this influence. As shown in Figure~\ref{fig:ablat_vp}, we find that removing PP results in a relatively larger TO value on the four datasets compared to that given by activating PP, this indicates that using PP helps reduce inter-class confusion between old classes. Concurrently,  we can also observe that the TN value obtained by removing PP is larger than that given by using PP, this suggests that using PP helps reduce inter-class confusion between new classes. As for FO and FN, we can see that using PP leads to a trade-off between FO and FN. For example, using PP increases/decreases the FN value but decreases/increases the FO value on FGVC-Aircraft/CIFAR100. Nonetheless, the positive effect (decreasing FO/FN) of using PP outweighs its negative effect (increasing FO/FN). In summary, \textit{rather than reducing the inter-class confusion between all classes, using PP mainly reduces the inter-class confusion between old/new classes.}

\begin{figure*}[t]
    \centering
    \includegraphics[width=2.0\columnwidth]{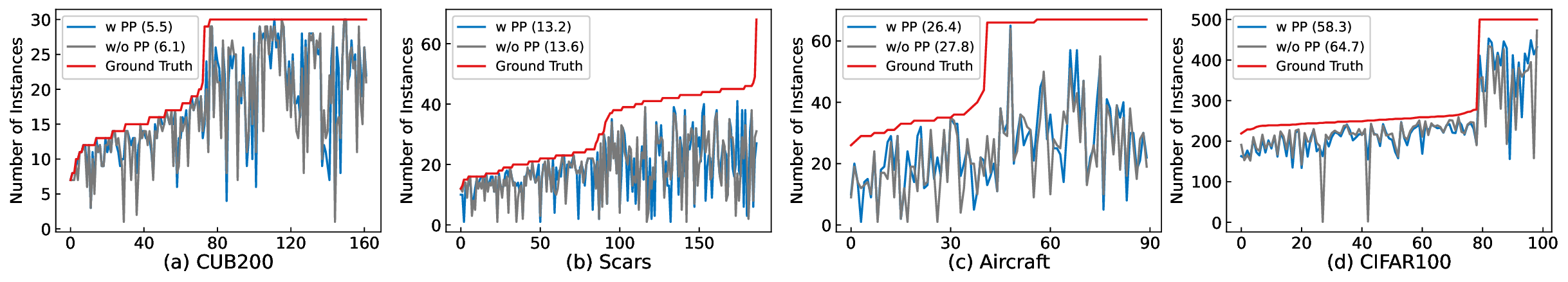}
    \caption{The number of instances contained in each class (red line) and correctly classified by deactivating (gray line) and activating (blue line) potential prototypes (PP), where the value shown in the bracket of the legend is the average bias given by the mean value of the difference between the number of correctly categorized instances and the actual number across all classes.}
    \label{fig:prediction_bias_across_on}
\end{figure*}

\begin{figure*}[h]
    \centering
    \includegraphics[width=1.93\columnwidth]{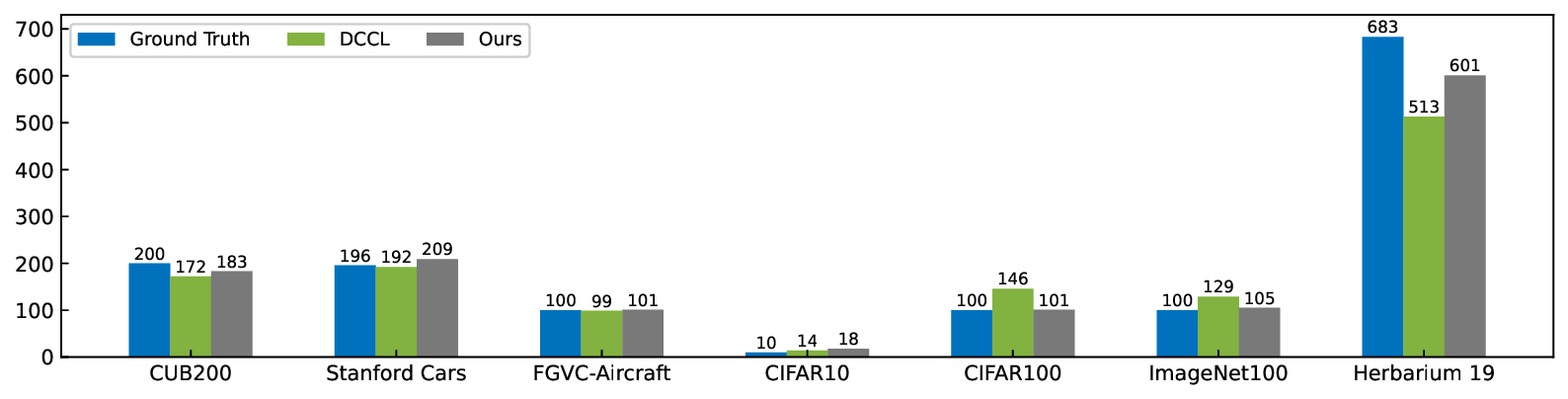}
    \caption{Category number of the unlabelled data given by the ground truth and estimated by DCCL and our proposed method.}
    \label{fig:num}
\end{figure*}

\subsubsection{Intra-Class Prediction Bias.}
To investigate whether using PP only reduces inter-class confusion, we conduct several experiments on CUB200, Stanford Cars, FGVC-Aircraft, and CIFAR100 to observe the intra-class prediction bias. Concretely, due to the same reason described in the preceding section, we first select classes that exist in both the clustering results given by removing and using PP. Then, we count the number of correctly recognized instances in each class and calculate the prediction bias by comparing that value with the ground truth. Finally, we average the prediction of each class to observe the overall performance. As shown in Figure~\ref{fig:prediction_bias_across_on}, we find that there exists a trade-off between the prediction bias in each class given by removing or using PP. Despite this, the average intra-class prediction bias given by removing PP is larger than that given by using PP, this indicates that using PP can also reduce the average intra-class prediction bias.

\subsubsection{Overlap ratio between $|\mathcal{Y}^{l}|$ and $|\mathcal{Y}^{u}|$.}

Considering more practical scenarios, $|\mathcal{Y}^{l}|$ and $|\mathcal{Y}^{u}|$ can be partially overlapped rather than just simple half-to-half. To validate the robustness of PNP to the number of overlapped classes, we vary the overlapping ratio from 25\% to 75\% and report the clustering accuracy in Table~\ref{tab:overlap}. Firstly, we can see that the clustering accuracy of each method shown in the table increases as the overlap ratio increases. Secondly, compared to the improvements achieved by GCD~\cite{vaze2022generalized} and GPC~\cite{zhao2023learning}, we find that the improvements achieved by our proposed PNP and DCCL~\cite{pu2023dynamic} are more significant. Finally, we observe that our proposed PNP consistently outperforms GCD~\cite{vaze2022generalized}, GPC~\cite{zhao2023learning}, and DCCL~\cite{pu2023dynamic} under the different overlap settings. In summary, the experimental results demonstrate the robustness and further validate our proposed PNP's effectiveness.

\begin{table}[t]\tiny
    \centering
    \begin{tabular}{cccccccccc}
    \toprule
    \multirow{2}{*}{Methods} & \multicolumn{3}{c}{25} & \multicolumn{3}{c}{50} & \multicolumn{3}{c}{75} \\
    \cmidrule(r){2-4}  \cmidrule(r){5-7} \cmidrule(r){8-10}
    {} & {All} & {Old} & {New}  & {All} & {Old} & {New} & {All} & {Old} & {New} \\
    \midrule
    
         GCD$^\dagger$~\cite{vaze2022generalized}     & 49.5 & 50.1 & 48.2 & 51.2 & 50.7 & 52.2 & 52.7 & 50.9 & 54.5 \\
         GPC~\cite{zhao2023learning}            & 51.2 & 52.6 & 49.5 & 52.3 & 51.6 & 54.8 & 53.6 & 51.4 & 55.9 \\
         DCCL$^*$~\cite{pu2023dynamic}          & {50.5} & {62.3} & {48.5} & {57.3} & {71.3} & {50.2} & {69.0} & {73.9} & {61.6} \\
         
    \midrule
         PNP(Ours)        & \textbf{54.0} & \textbf{64.5} & \textbf{52.3} & \textbf{62.2} & \textbf{75.4} & \textbf{55.6} & \textbf{72.4} & \textbf{75.4} & \textbf{67.9} \\ 
   \bottomrule  
    \end{tabular}
    \caption{Results of different overlap between $|\mathcal{Y}^{l}|$ and $|\mathcal{Y}^{u}|$ on CUB200, where $^\dagger$ refers to results reported in GPC, $^*$ represents our reproduced results, and the overlap ration is computed by $100\times\frac{|\mathcal{Y}^{l}|}{|\mathcal{Y}^{u}|}$.}
    \label{tab:overlap}
\end{table}

\subsubsection{Limitations.}

As shown in Figure~\ref{fig:num}, PNP prefers to estimate a relatively larger category number compared to that estimated by DCCL~\cite{pu2023dynamic} in most of the datasets. The advantage of this trait is that when the category number in the dataset is large (\eg, CUB200), our proposed method can give a category number estimation result that is relatively closer to the ground truth. The disadvantage is that when the category number in the dataset is small (e.g., CIFAR10), the category number estimated by our proposed method has a large deviation from the ground truth, thus affecting the performance seriously. Moreover, we find our proposed PNP is susceptible to the number of neighbors in the clustering process. We leave how to adaptively obtain a suitable number of neighbors based on the characteristics of different datasets as an open-discussed question.

\section{Conclusion}\label{sec:conclusion and future work}

In this paper, we propose a straightforward yet efficacious method for addressing the complex task of GCD. Specifically, our proposed method, dubbed Probing New Prototype (PNP), adopts an end-to-end manner that alternates between estimating the number of categories and learning their representations. Notably, PNP employs potential prototypes to facilitate the division of clusters containing instances from divergent classes. This is achieved through the activation of potential prototypes via a teacher prober, which produces soft predictions. Despite its simplicity, PNP demonstrates superior performance, setting a new benchmark for state-of-the-art among one-stage GCD methods.

\section*{Acknowledgements}
This work was supported by in part by the NSFC under Grant 62272380 and 62103317, the Science and Technology Program of Xi’an,
China under Grant 21RGZN0017, Natural Science Foundation in Shaanxi Province of China under grant 2021JQ-289, and China Postdoctoral Science Foundation under Grant 2021M700533.

{\small
\bibliographystyle{ieee_fullname}
\bibliography{ref}

\begin{thebibliography}{10}\itemsep=-1pt

\bibitem{an2024transfer}
Wenbin An, Feng Tian, Wenkai Shi, Yan Chen, Yaqiang Wu, Qianying Wang, and Ping
  Chen.
\newblock Transfer and alignment network for generalized category discovery.
\newblock In {\em Proceedings of the AAAI Conference on Artificial
  Intelligence}, volume~38, pages 10856--10864, 2024.

\bibitem{an2023generalized}
Wenbin An, Feng Tian, Qinghua Zheng, Wei Ding, QianYing Wang, and Ping Chen.
\newblock Generalized category discovery with decoupled prototypical network.
\newblock In {\em Proceedings of the AAAI Conference on Artificial
  Intelligence}, volume~37, pages 12527--12535, 2023.

\bibitem{assran2022masked}
Mahmoud Assran, Mathilde Caron, Ishan Misra, Piotr Bojanowski, Florian Bordes,
  Pascal Vincent, Armand Joulin, Mike Rabbat, and Nicolas Ballas.
\newblock Masked siamese networks for label-efficient learning.
\newblock In {\em European Conference on Computer Vision}, pages 456--473.
  Springer, 2022.

\bibitem{assran2021semi}
Mahmoud Assran, Mathilde Caron, Ishan Misra, Piotr Bojanowski, Armand Joulin,
  Nicolas Ballas, and Michael Rabbat.
\newblock Semi-supervised learning of visual features by non-parametrically
  predicting view assignments with support samples.
\newblock In {\em Proceedings of the IEEE/CVF International Conference on
  Computer Vision}, pages 8443--8452, 2021.

\bibitem{berthelot2019mixmatch}
David Berthelot, Nicholas Carlini, Ian Goodfellow, Nicolas Papernot, Avital
  Oliver, and Colin~A Raffel.
\newblock Mixmatch: A holistic approach to semi-supervised learning.
\newblock {\em Advances in neural information processing systems}, 32, 2019.

\bibitem{cao2021openworld}
Kaidi Cao, Maria Brbic, and Jure Leskovec.
\newblock Open-world semi-supervised learning, 2021.

\bibitem{caron2020unsupervised}
Mathilde Caron, Ishan Misra, Julien Mairal, Priya Goyal, Piotr Bojanowski, and
  Armand Joulin.
\newblock Unsupervised learning of visual features by contrasting cluster
  assignments.
\newblock {\em Advances in neural information processing systems},
  33:9912--9924, 2020.

\bibitem{caron2021emerging}
Mathilde Caron, Hugo Touvron, Ishan Misra, Herv{\'e} J{\'e}gou, Julien Mairal,
  Piotr Bojanowski, and Armand Joulin.
\newblock Emerging properties in self-supervised vision transformers.
\newblock In {\em Proceedings of the IEEE/CVF international conference on
  computer vision}, pages 9650--9660, 2021.

\bibitem{chen2020simple}
Ting Chen, Simon Kornblith, Mohammad Norouzi, and Geoffrey Hinton.
\newblock A simple framework for contrastive learning of visual
  representations.
\newblock In {\em International conference on machine learning}, pages
  1597--1607. PMLR, 2020.

\bibitem{deng2009imagenet}
Jia Deng, Wei Dong, Richard Socher, Li-Jia Li, Kai Li, and Li Fei-Fei.
\newblock Imagenet: A large-scale hierarchical image database.
\newblock In {\em 2009 IEEE conference on computer vision and pattern
  recognition}, pages 248--255. Ieee, 2009.

\bibitem{dosovitskiy2020image}
Alexey Dosovitskiy, Lucas Beyer, Alexander Kolesnikov, Dirk Weissenborn,
  Xiaohua Zhai, Thomas Unterthiner, Mostafa Dehghani, Matthias Minderer, Georg
  Heigold, Sylvain Gelly, Jakob Uszkoreit, and Neil Houlsby.
\newblock An image is worth 16x16 words: Transformers for image recognition at
  scale.
\newblock In {\em ICLR}, 2021.

\bibitem{fini2021unified}
Enrico Fini, Enver Sangineto, St{\'e}phane Lathuili{\`e}re, Zhun Zhong, Moin
  Nabi, and Elisa Ricci.
\newblock A unified objective for novel class discovery.
\newblock In {\em Proceedings of the IEEE/CVF International Conference on
  Computer Vision}, pages 9284--9292, 2021.

\bibitem{girish2021towards}
Sharath Girish, Saksham Suri, Sai~Saketh Rambhatla, and Abhinav Shrivastava.
\newblock Towards discovery and attribution of open-world gan generated images.
\newblock In {\em Proceedings of the IEEE/CVF International Conference on
  Computer Vision}, pages 14094--14103, 2021.

\bibitem{grill2020bootstrap}
Jean-Bastien Grill, Florian Strub, Florent Altch{\'e}, Corentin Tallec, Pierre
  Richemond, Elena Buchatskaya, Carl Doersch, Bernardo Avila~Pires, Zhaohan
  Guo, Mohammad Gheshlaghi~Azar, et~al.
\newblock Bootstrap your own latent-a new approach to self-supervised learning.
\newblock {\em Advances in neural information processing systems},
  33:21271--21284, 2020.

\bibitem{han2021autonovel}
Kai Han, Sylvestre-Alvise Rebuffi, Sebastien Ehrhardt, Andrea Vedaldi, and
  Andrew Zisserman.
\newblock Autonovel: Automatically discovering and learning novel visual
  categories.
\newblock {\em IEEE Transactions on Pattern Analysis and Machine Intelligence},
  44(10):6767--6781, 2021.

\bibitem{han2019learning}
Kai Han, Andrea Vedaldi, and Andrew Zisserman.
\newblock Learning to discover novel visual categories via deep transfer
  clustering.
\newblock In {\em Proceedings of the IEEE/CVF International Conference on
  Computer Vision}, pages 8401--8409, 2019.

\bibitem{hao2024cipr}
Shaozhe Hao, Kai Han, and Kwan-Yee~K. Wong.
\newblock Ci{PR}: An efficient framework with cross-instance positive relations
  for generalized category discovery.
\newblock {\em Transactions on Machine Learning Research}, 2024.

\bibitem{he2020momentum}
Kaiming He, Haoqi Fan, Yuxin Wu, Saining Xie, and Ross Girshick.
\newblock Momentum contrast for unsupervised visual representation learning.
\newblock In {\em Proceedings of the IEEE/CVF conference on computer vision and
  pattern recognition}, pages 9729--9738, 2020.

\bibitem{ResNet}
Kaiming He, Xiangyu Zhang, Shaoqing Ren, and Jian Sun.
\newblock Deep residual learning for image recognition.
\newblock In {\em CVPR}, pages 770--778, 2016.

\bibitem{hsu2018learning}
Yen-Chang Hsu, Zhaoyang Lv, and Zsolt Kira.
\newblock Learning to cluster in order to transfer across domains and tasks.
\newblock In {\em International Conference on Learning Representations}, 2018.

\bibitem{hsu2019multi}
Yen-Chang Hsu, Zhaoyang Lv, Joel Schlosser, Phillip Odom, and Zsolt Kira.
\newblock Multi-class classification without multi-class labels.
\newblock {\em arXiv preprint arXiv:1901.00544}, 2019.

\bibitem{huang2021deep}
Jiabo Huang and Shaogang Gong.
\newblock Deep clustering by semantic contrastive learning.
\newblock {\em arXiv preprint arXiv:2103.02662}, 2021.

\bibitem{jia2021joint}
Xuhui Jia, Kai Han, Yukun Zhu, and Bradley Green.
\newblock Joint representation learning and novel category discovery on
  single-and multi-modal data.
\newblock In {\em Proceedings of the IEEE/CVF International Conference on
  Computer Vision}, pages 610--619, 2021.

\bibitem{johnson2019billion}
Jeff Johnson, Matthijs Douze, and Herv{\'e} J{\'e}gou.
\newblock Billion-scale similarity search with gpus.
\newblock {\em IEEE Transactions on Big Data}, 7(3):535--547, 2019.

\bibitem{kim2023proxy}
Hyungmin Kim, Sungho Suh, Daehwan Kim, Daun Jeong, Hansang Cho, and Junmo Kim.
\newblock Proxy anchor-based unsupervised learning for continuous generalized
  category discovery.
\newblock In {\em Proceedings of the IEEE/CVF International Conference on
  Computer Vision}, pages 16688--16697, 2023.

\bibitem{krause20133d}
Jonathan Krause, Michael Stark, Jia Deng, and Li Fei-Fei.
\newblock 3d object representations for fine-grained categorization.
\newblock In {\em Proceedings of the IEEE international conference on computer
  vision workshops}, pages 554--561, 2013.

\bibitem{krizhevsky2009learning}
Alex Krizhevsky, Geoffrey Hinton, et~al.
\newblock Learning multiple layers of features from tiny images.
\newblock 2009.

\bibitem{li2020prototypical}
Junnan Li, Pan Zhou, Caiming Xiong, and Steven~CH Hoi.
\newblock Prototypical contrastive learning of unsupervised representations.
\newblock {\em arXiv preprint arXiv:2005.04966}, 2020.

\bibitem{loshchilov2017sgdr}
Ilya Loshchilov and Frank Hutter.
\newblock {SGDR}: Stochastic gradient descent with warm restarts.
\newblock In {\em International Conference on Learning Representations}, 2017.

\bibitem{ma2024active}
Shijie Ma, Fei Zhu, Zhun Zhong, Xu-Yao Zhang, and Cheng-Lin Liu.
\newblock Active generalized category discovery.
\newblock {\em arXiv preprint arXiv:2403.04272}, 2024.

\bibitem{macqueen1967some}
James MacQueen et~al.
\newblock Some methods for classification and analysis of multivariate
  observations.
\newblock In {\em Proceedings of the fifth Berkeley symposium on mathematical
  statistics and probability}, volume~1, pages 281--297. Oakland, CA, USA,
  1967.

\bibitem{maji2013fine}
Subhransu Maji, Esa Rahtu, Juho Kannala, Matthew Blaschko, and Andrea Vedaldi.
\newblock Fine-grained visual classification of aircraft.
\newblock {\em arXiv preprint arXiv:1306.5151}, 2013.

\bibitem{learning2006semi}
Bernhard~Scholkopf Olivier~Chapelle and Alexander Zien.
\newblock Semi-supervised learning.
\newblock {\em MIT Press}, 2006.

\bibitem{parkhi2012cats}
Omkar~M Parkhi, Andrea Vedaldi, Andrew Zisserman, and CV Jawahar.
\newblock Cats and dogs.
\newblock In {\em 2012 IEEE conference on computer vision and pattern
  recognition}, pages 3498--3505. IEEE, 2012.

\bibitem{pu2023dynamic}
Nan Pu, Zhun Zhong, and Nicu Sebe.
\newblock Dynamic conceptional contrastive learning for generalized category
  discovery.
\newblock In {\em Proceedings of the IEEE/CVF Conference on Computer Vision and
  Pattern Recognition}, pages 7579--7588, 2023.

\bibitem{rastegar2023learn}
Sarah Rastegar, Hazel Doughty, and Cees Snoek.
\newblock Learn to categorize or categorize to learn? self-coding for
  generalized category discovery.
\newblock In {\em Thirty-seventh Conference on Neural Information Processing
  Systems}, 2023.

\bibitem{rosvall2008maps}
Martin Rosvall and Carl~T Bergstrom.
\newblock Maps of random walks on complex networks reveal community structure.
\newblock {\em Proceedings of the national academy of sciences},
  105(4):1118--1123, 2008.

\bibitem{tan2019herbarium}
Kiat~Chuan Tan, Yulong Liu, Barbara Ambrose, Melissa Tulig, and Serge Belongie.
\newblock The herbarium challenge 2019 dataset.
\newblock {\em arXiv preprint arXiv:1906.05372}, 2019.

\bibitem{van2008visualizing}
Laurens Van~der Maaten and Geoffrey Hinton.
\newblock Visualizing data using t-sne.
\newblock {\em Journal of machine learning research}, 9(11), 2008.

\bibitem{vaze2022generalized}
Sagar Vaze, Kai Han, Andrea Vedaldi, and Andrew Zisserman.
\newblock Generalized category discovery.
\newblock In {\em Proceedings of the IEEE/CVF Conference on Computer Vision and
  Pattern Recognition}, pages 7492--7501, 2022.

\bibitem{vaze2022openset}
Sagar Vaze, Kai Han, Andrea Vedaldi, and Andrew Zisserman.
\newblock Open-set recognition: a good closed-set classifier is all you need?
\newblock In {\em International Conference on Learning Representations}, 2022.

\bibitem{vaze2023no}
Sagar Vaze, Andrea Vedaldi, and Andrew Zisserman.
\newblock No representation rules them all in category discovery.
\newblock In {\em Thirty-seventh Conference on Neural Information Processing
  Systems}, 2023.

\bibitem{cub}
Catherine Wah, Steve Branson, Peter Welinder, Pietro Perona, and Serge
  Belongie.
\newblock The caltech-ucsd birds-200-2011 dataset.
\newblock {\em california institute of technology}, 2011.

\bibitem{wang2024sptnet}
Hongjun Wang, Sagar Vaze, and Kai Han.
\newblock {SPTN}et: An efficient alternative framework for generalized category
  discovery with spatial prompt tuning.
\newblock In {\em The Twelfth International Conference on Learning
  Representations}, 2024.

\bibitem{wen2023parametric}
Xin Wen, Bingchen Zhao, and Xiaojuan Qi.
\newblock Parametric classification for generalized category discovery: A
  baseline study.
\newblock In {\em Proceedings of the IEEE/CVF International Conference on
  Computer Vision}, pages 16590--16600, 2023.

\bibitem{wu2023metagcd}
Yanan Wu, Zhixiang Chi, Yang Wang, and Songhe Feng.
\newblock Metagcd: Learning to continually learn in generalized category
  discovery.
\newblock In {\em Proceedings of the IEEE/CVF International Conference on
  Computer Vision}, pages 1655--1665, 2023.

\bibitem{yang2023generalized}
Xiangli Yang, Xinglin Pan, Irwin King, and Zenglin Xu.
\newblock Generalized category discovery with clustering assignment
  consistency.
\newblock In {\em International Conference on Neural Information Processing},
  pages 535--547. Springer, 2023.

\bibitem{zhang2023promptcal}
Sheng Zhang, Salman Khan, Zhiqiang Shen, Muzammal Naseer, Guangyi Chen, and
  Fahad~Shahbaz Khan.
\newblock Promptcal: Contrastive affinity learning via auxiliary prompts for
  generalized novel category discovery.
\newblock In {\em Proceedings of the IEEE/CVF Conference on Computer Vision and
  Pattern Recognition}, pages 3479--3488, 2023.

\bibitem{zhao2021novel}
Bingchen Zhao and Kai Han.
\newblock Novel visual category discovery with dual ranking statistics and
  mutual knowledge distillation.
\newblock {\em Advances in Neural Information Processing Systems},
  34:22982--22994, 2021.

\bibitem{zhao2023incremental}
Bingchen Zhao and Oisin Mac~Aodha.
\newblock Incremental generalized category discovery.
\newblock In {\em Proceedings of the IEEE/CVF International Conference on
  Computer Vision}, pages 19137--19147, 2023.

\bibitem{zhao2023learning}
Bingchen Zhao, Xin Wen, and Kai Han.
\newblock Learning semi-supervised gaussian mixture models for generalized
  category discovery.
\newblock {\em arXiv preprint arXiv:2305.06144}, 2023.

\bibitem{zhong2021neighborhood}
Zhun Zhong, Enrico Fini, Subhankar Roy, Zhiming Luo, Elisa Ricci, and Nicu
  Sebe.
\newblock Neighborhood contrastive learning for novel class discovery.
\newblock In {\em Proceedings of the IEEE/CVF conference on computer vision and
  pattern recognition}, pages 10867--10875, 2021.

\end{thebibliography}
}
\clearpage

\appendix




\section{Losses}

\begin{table}[ht]\small
    \begin{center}
    \begin{tabular}{c|c|ccc|ccc}
    \toprule
       \multirow{2}{*}{Exp.} &  \multirow{2}{*}{Method}  & \multicolumn{3}{c|}{CUB200} & \multicolumn{3}{c}{CIFAR100} \\
    \cmidrule{3-8}
         {} & {}    & {All} & {Old} & {New}  & {All} & {Old} & {New}\\
    \midrule
         {1)} & {w/o $\mathcal{L}_{\text{crl}}$}     & {55.0}  & {46.2} & {59.0} & {74.6}  & {77.4} & {69.0} \\
         {2)} & {w/o $\mathcal{L}_{\text{cru}}$}     & {60.0}  & {59.3} & {60.3} & {58.1}  & {69.8} & {34.2} \\
         {3)} & {w/o $\mathcal{L}_{\text{sup}}$}     & {60.1}  & {55.3} & {62.5} & {72.4}  & {75.2} & {66.8} \\
         {4)} & {w/o $\mathcal{L}_{\text{unsup}}$}   & {63.1}  & {61.7} & {63.8} & {79.7}  & {80.8} & \textbf{77.5} \\
    \midrule
         {5)} & {w all}                              & \textbf{67.4}  & \textbf{69.2} & \textbf{66.5} & \textbf{80.0}  & \textbf{81.7} & {76.3} \\
    \bottomrule
    \end{tabular}
    \caption{Ablation studies on losses.}
    \label{tab:ablat_losses}
    \end{center}
\end{table}

To investigate the influence of each loss used in our proposed PNP, we remove each loss individually and compare the corresponding clustering accuracy with that given by using all losses. As shown in \cref{tab:ablat_losses}, removing each loss drops the clustering accuracy indicating that each loss is important for PNP. Further, comparing the clustering accuracy given by removing other losses, removing $\mathcal{L}_{\text{crl}}$ drops the clustering accuracy by a relatively larger margin on CUB200. Conversely, removing $\mathcal{L}_{\text{cru}}$ results in a larger drop in the clustering accuracy compared to that given by removing other losses on CIFAR100. The reason may stem from that CUB200 is a fine-grained dataset that needs stronger supervision compared to the coarse-grained dataset CIFAR100 to split each class well.

\section{Trainable \textit{VS} Frozen}
\begin{table}[h]
    \centering
    \begin{tabular}{l|cccc}
    \toprule
        Method & {$K^e$} & All & Old & New \\
    \midrule
        w/o PP         &  175 & 62.4 & 59.1 & 64.1 \\ 
        Frozen  PP     &  183 & 66.4 & 65.6 & 66.9 \\
        Trainable PP   &  183 & 67.4 & 69.2 & 66.5 \\
    \bottomrule
    \end{tabular}
    \caption{Estimated category number and clustering accuracy of different settings, where PP refers to the potential prototypes. }
    \label{tab:fvt}
\end{table}

\begin{table*}[ht]\small
    \centering
    \begin{tabular}{c|c|ccc|ccc|ccc|ccc}
    \toprule
    \multirow{2}{*}{Methods}& \multirow{2}{*}{Known $K^u$} & \multicolumn{3}{c|}{CUB200} & \multicolumn{3}{c|}{Stanford Cars} & \multicolumn{3}{c}{FGVC-Aircraft} & \multicolumn{3}{c}{Oxford-Pet}\\
    \cmidrule(r){3-5}  \cmidrule(r){6-8} \cmidrule(r){9-11} \cmidrule(r){12-14}
    {} & {} & {All} & {Old} & {New}  & {All} & {Old} & {New} & {All} & {Old} & {New} & {All} & {Old} & {New} \\
    \midrule
         k-means~\cite{macqueen1967some}    & \ding{51}  & 34.3 & 38.9 & 32.1 & 12.8 & 10.6 & 13.8 & 16.0 & 14.4 & 16.8 & {77.1}  & {70.1} & {80.7} \\
         RS+~\cite{han2021autonovel}        & \ding{51}  & 33.3 & 51.6 & 24.2 & 28.3 & 61.8 & 12.1 & 26.9 & 36.4 & 22.2 & {-}  & {-} & {-}\\
         UNO+~\cite{fini2021unified}        & \ding{51}  & 35.1 & 49.0 & 28.1 & 35.5 & 70.5 & 18.6 & 40.3 & 56.4 & 32.2 & {-}  & {-} & {-} \\    
         ORCA~\cite{cao2021openworld}       & \ding{51}  & 35.3 & 45.6 & 30.2 & 23.5 & 50.1 & 10.7 & 22.0 & 31.8 & 17.1 & {-}  & {-} & {-} \\
         GCD~\cite{vaze2022generalized}     & \ding{51}  & 51.3 & 56.6 & 48.7 & 39.0 & 57.6 & 29.9 & 45.0 & 41.1 & 46.9 & {-}  & {-} & {-}\\
         CiPR~\cite{hao2024cipr}            & \ding{51}  & 57.1 & 58.7 & 55.6 & 47.0 & 61.5 & 40.1 & {-}  & {-} & {-} & {-}  & {-} & {-} \\ 
         SimGCD~\cite{wen2023parametric}    & \ding{51}  & 60.3 & 65.6 & 57.7 & 53.8 & 71.9 & 45.0 & 54.2 & 59.1 & 51.8 & {-}  & {-} & {-} \\
         PromptCAL~\cite{zhang2023promptcal}& \ding{51}  & 62.9 & 64.4 & 62.1 & 50.2 & 70.1 & 40.6 & 52.2 & 52.2 & 52.3 & {-}  & {-} & {-} \\
         $\mu$GCD~\cite{vaze2023no}         & \ding{51}  & 65.7 & 68.0 & {64.6} & {56.5} & 68.1 & \textbf{50.9} & 53.8 & 55.4 & {53.0} & {-}  & {-} & {-} \\
         SPTNet~\cite{wang2024sptnet} & \ding{51}  & 65.8 & 68.8 & {65.1} & \textbf{59.0} & \textbf{79.2} & 49.3 & \textbf{59.3} & 61.8 & \textbf{58.1} & {-}  & {-} & {-}\\
         InfoSieve~\cite{rastegar2023learn} & \ding{51}  & \textbf{69.4} & \textbf{77.9} & \textbf{65.2} & {55.7} & {74.8} & 46.4 & {56.3} & \textbf{63.7} & 52.5 & \textbf{91.8}  & \textbf{92.6} & \textbf{91.3}\\

         \midrule
         GCD$^*$~\cite{vaze2022generalized} & \ding{55}   & 47.1 & 55.1 & 44.8 & 35.0 & 56.0 & 24.8 & 40.1 & 40.8 & 42.8 & {80.2}  & {85.1} & {77.6} \\
         Yang \etal~\cite{yang2023generalized} & \ding{55}& 58.0 & 65.0 & 43.9 & 47.6 & \textbf{70.6} & 33.8 & {-}  & {-} & {-} & {-}  & {-} & {-}\\ 
         GPC~\cite{zhao2023learning}        & \ding{55}   & 52.0 & 55.5 & 47.5 & 38.2 & 58.9 & 27.4 & 43.3 & 40.7 & 44.8 & {-}  & {-} & {-}\\
         DCCL$^*$~\cite{pu2023dynamic}      & \ding{55}   & 63.5 & 60.8 & 64.9 & 43.1 & 55.7 & 36.2 & 42.7 & 41.5 & 43.4 & {88.1}  & \textbf{88.2} & {88.0}\\
    
         PNP(Ours)        & \ding{55} & \textbf{67.4} & \textbf{69.2} & \textbf{66.5} & \textbf{57.3} & {70.2} & \textbf{51.1} & \textbf{47.0} & \textbf{49.5} & {45.7} & \textbf{89.7}  & {87.6} & \textbf{90.8}\\
         
   \bottomrule  
    \end{tabular}
    \caption{Results on the Semantic Shift Benchmark, where $^*$ denotes our reproduced results or from their publicly available results, $-$ denotes no results reported in the corresponding works, and best results with/without taking the category number $K^u$ as the prior are highlighted in bold font.}
    \label{tab:ssb_new}
\end{table*}


\begin{table*}[ht]\small
    \centering
    \begin{tabular}{c|c|ccc|ccc|ccc|ccc}
    \toprule
    \multirow{2}{*}{Methods}& \multirow{2}{*}{Known $K^u$} & \multicolumn{3}{c|}{CIFAR10} & \multicolumn{3}{c|}{CIFAR100} & \multicolumn{3}{c|}{ImageNet100} & \multicolumn{3}{c}{Herbarium 19} \\
    \cmidrule(r){3-5}  \cmidrule(r){6-8} \cmidrule(r){9-11} \cmidrule(r){12-14}
    {} & {} & {All} & {Old} & {New}  & {All} & {Old} & {New} & {All} & {Old} & {New} & {All} & {Old} & {New} \\
    \midrule
         k-means~\cite{macqueen1967some}    & \ding{51} & 83.6 & 85.7 & 82.5 & 52.0 & 52.2 & 50.8 & 72.7 & 75.5 & 71.3 & 13.0 & 12.2 & 13.4 \\
         RankStats+~\cite{han2021autonovel} & \ding{51} & 46.8 & 19.2 & 60.5 & 58.2 & 77.6 & 19.3 & 37.1 & 61.6 & 24.8 & 27.9 & 55.8 & 12.8 \\
         UNO+~\cite{fini2021unified}        & \ding{51} & 68.6 & 98.3 & 53.8 & 69.5 & 80.6 & 47.2 & 70.3 & 95.0 & 57.9 & 28.3 & 53.7 & 14.7 \\   
         ORCA~\cite{cao2021openworld}       & \ding{51} & 81.8 & 86.2 & 79.6 & 69.0 & 77.4 & 52.0 & 73.5 & 92.6 & 63.9 & 20.9 & 30.9 & 15.5 \\
         GCD~\cite{vaze2022generalized}     & \ding{51} & 91.5 & \textbf{97.9} & 88.2 & 73.0 & 76.2 & 66.5 & {72.7}  & {91.8} & {63.8} & 35.4 & 51.0 & 27.0 \\
         SimGCD~\cite{wen2023parametric}    & \ding{51}  & 97.1 & 95.1 & {98.1} & 80.1 & 81.2 & {77.8} & {83.0} & {93.1} & {77.9} & \textbf{44.0} & 58.0 & \textbf{36.4} \\
         InfoSieve~\cite{rastegar2023learn} & \ding{51}  & 94.8 & 97.7 & 93.4  & 78.3 & 82.2 & 70.5 & {80.5} & {93.8} & {73.8} & 41.0 & \textbf{55.4} & 33.2 \\
         CiPR~\cite{hao2024cipr}            & \ding{51}  & 97.7 & 97.5 & 97.7 & \textbf{81.5} & 82.4 & \textbf{79.7} & 80.5 & 84.9 & 78.3 & 36.8 & 45.4 & 32.6\\
         SPTNet~\cite{wang2024sptnet} & \ding{51}  & 97.3 & 95.0 & \textbf{98.6} & 81.3 & \textbf{84.3} & 75.6 & \textbf{85.4} & \textbf{93.2} & \textbf{81.4} & 43.4 & 58.7 & 35.2 \\
         PromptCAL~\cite{zhang2023promptcal}& \ding{51}  & \textbf{97.9} & {96.6} & {98.5} & {81.2} & {84.2} & 75.3 & {83.1} & {92.7} & {78.3} & - & - & - \\

         \midrule
         GCD$^*$~\cite{vaze2022generalized}& \ding{55}   & 88.6 & 96.2 & 84.9 & 73.0 & 76.2 & 66.5 & {72.7} & {91.8} & {63.8} & 30.9 & 36.3 & 28.0 \\
         GPC~\cite{zhao2023learning}       & \ding{55}   & 90.6 & \textbf{97.6} & 87.0 & 75.4 & \textbf{84.6} & 60.1 & {75.3} & \textbf{93.4} & {66.7} & 36.5 & 51.7 & 27.9 \\
         Yang \etal~\cite{yang2023generalized} & \ding{55}& 92.3 & 91.4 & 94.4 & 78.5 & 81.4 & 75.6 & 81.1 & 80.3 & \textbf{81.8} & 36.3 & 53.1 & 30.7 \\
         DCCL$^*$~\cite{pu2023dynamic}     & \ding{55}   & \textbf{96.3} & {96.5} & \textbf{96.9} & 75.3 & 76.8 & 70.2 & {80.5} & {90.5} & {76.2} & 39.8 & 50.5 & \textbf{34.1} \\
         PNP(Ours)                         & \ding{55}   & {90.2} & {95.0} & {87.8} & \textbf{80.0} & {80.7} & \textbf{76.3} & \textbf{82.4} & {92.9} & {77.2} & \textbf{41.1} & \textbf{56.2} & {32.9} \\
         
   \bottomrule  
    \end{tabular}
    \caption{Results on generic image recognition datasets and more challenging fine-grained and long-tailed datasets, where $^*$ denotes our reproduced results or from their publicly available results. Best results with/without taking the category number $K^u$ as the prior are highlighted in bold font. }
    \label{tab:cch_new}
\end{table*}

\begin{table*}[ht]\small
    \centering
    \begin{tabular}{c|c|ccc|ccc|ccc|ccc}
    \toprule
    \multirow{2}{*}{Methods}& \multirow{2}{*}{Known $K^u$} & \multicolumn{3}{c|}{Herbarium 19} & \multicolumn{3}{c|}{Stanford Cars} & \multicolumn{3}{c}{FGVC-Aircraft} & \multicolumn{3}{c}{CIFAR10}\\
    \cmidrule(r){3-5}  \cmidrule(r){6-8} \cmidrule(r){9-11} \cmidrule(r){12-14}
    {} & {} & {All} & {Old} & {New}  & {All} & {Old} & {New} & {All} & {Old} & {New} & {All} & {Old} & {New} \\
    \midrule
         SimGCD~\cite{wen2023parametric}     & \ding{51}  & 44.0 & 58.0 & 36.4 & 53.8 & 71.9 & 45.0 & 54.2 & 59.1 & 51.8 & 97.1 & 95.1 & 98.1 \\
         SimGCD$^*$~\cite{wen2023parametric} & \ding{55}  & 39.7 & 58.2 & 29.7 & 49.1 & 65.1 & 41.3 & 43.7 & 33.1 & 47.1 & 89.5 & 86.5 & 91.1 \\

         PNP(Ours)                           & \ding{55}  & 41.1 & 56.2 & 32.9 & 57.3 & 70.2 & 51.1 & 47.0 & 49.5 & 45.7 & 90.2 & 95.0 & 87.8\\
         
   \bottomrule  
    \end{tabular}
    \caption{Clustering accuracy of SimGCD and our proposed PNP, where $^*$ denotes our reproduced results.}
    \label{tab:ssb_new_new}
\end{table*}

Table \ref{tab:fvt} shows the estimated category number and clustering accuracy given by removing potential prototypes (PP), freezing PP, and setting PP trainable, respectively. We can see that compared to the performance given by removing PP, using frozen PP or trainable PP both increases the estimated category number and improves the clustering accuracy. The results further validate that using PP helps discover more novel classes. Compared to the clustering accuracy given by using frozen PP, though using trainable PP drops the performance on ``NEW'' classes slightly, it improves the performance on ``OLD'' classes by a relatively larger margin. This indicates that using trainable PP is helpful.

\section{More Comparisons.}

Table~\ref{tab:ssb_new} and table\ref{tab:cch_new} shows the clustering accuracy of other methods adapted to solve the GCD task,
including the conventional clustering algorithm (k-means~\cite{macqueen1967some}), NCD methods (RankStats+~\cite{han2021autonovel} and UNO+~\cite{fini2021unified}), and method used to solve the open-world semi-supervised learning (ORCA~\cite{cao2021openworld}). From this table, we can see that though these methods take the category number as a prior, our proposed PNP still consistently outperforms these methods.

Table~\ref{tab:ssb_new} and table\ref{tab:cch_new} show the publicly available clustering accuracy of some methods on {\textbf{Oxford Pets}}~\cite{parkhi2012cats}. The Oxford Pets is a fine-grained dataset that consists of 3.6k instances from 37 classes.  Align with~\cite{rastegar2023learn,pu2023dynamic}, we sample 19 classes as the  ``Old'' classes ($\mathcal{Y}^l$) and and keep the rest as ``New'' ($\mathcal{Y}^u \textbackslash \mathcal{Y}^l$). Compared to DCCL~\cite{pu2023dynamic},  higher clustering accuracy on ``All'' classes is achieved by our proposed PNP demonstrating that PNP is also more effective than DCCL on the  Oxford Pets.

In methods that take the category number as a prior, SimGCD~\cite{wen2023parametric} is most similar to our proposed PNP, we reproduce SimGCD by eliminating the assumption of a known category number and report corresponding results in \cref{tab:ssb_new_new}. We can see that eliminating the prior drops the performance of SimGCD significantly. Moreover, under the original setting of GCD, \ie, the category number of unlabelled data is unknown, our proposed PNP consistently outperforms SimGCD.

\end{document}